\documentclass[letterpaper]{article} 
\usepackage{aaai2026}  
\usepackage{times}  
\usepackage{helvet}  
\usepackage{courier}  
\usepackage[hyphens]{url}  
\usepackage{graphicx} 
\usepackage{natbib}  
\usepackage{caption} 
\usepackage{subcaption} 

\usepackage{algorithm}
\usepackage{algpseudocode}

\usepackage{newfloat}
\usepackage{listings}
\DeclareCaptionStyle{ruled}{labelfont=normalfont,labelsep=colon,strut=off} 
\floatstyle{ruled}
\newfloat{listing}{tb}{lst}{}
\floatname{listing}{Listing}
\usepackage{multirow}
\usepackage{amsmath}
\usepackage{url}
\usepackage{booktabs}
\usepackage{amssymb}
\usepackage{soul} 
\usepackage{pifont}
\usepackage[table]{xcolor}

\title{Multi-Stage Knowledge-Distilled VGAE and GAT for Robust Controller-Area-Network Intrusion Detection\\
}
\author{
Robert Frenken, Sidra Ghayour Bhatti, Hanqin Zhang, Qadeer Ahmed 
}
\affiliations {
The Ohio State University, Columbus, OH, USA\\
\{frenken.2, bhatti.39, zhang.14641, ahmed.358\}@osu.edu
}
\nocopyright
\begin{document}

\maketitle

\begin{abstract}
The Controller Area Network (CAN) protocol is a standard for in-vehicle communication but remains susceptible to cyber-attacks due to its lack of built-in security. This paper presents a multi-stage intrusion detection framework leveraging unsupervised anomaly detection and supervised graph learning tailored for automotive CAN traffic. Our architecture combines a Variational Graph Autoencoder (VGAE) for structural anomaly detection with a Knowledge-Distilled Graph Attention Network (KD-GAT) for robust attack classification. CAN bus activity is encoded as graph sequences to model temporal and relational dependencies. The pipeline applies VGAE-based selective undersampling to address class imbalance, followed by GAT classification with optional score-level fusion. The compact student GAT achieves 96\% parameter reduction compared to the teacher model while maintaining strong predictive performance. Experiments on six public CAN intrusion datasets—Car-Hacking, Car-Survival, and can-train-and-test—demonstrate competitive accuracy and efficiency, with average improvements of 16.2\% in F1-score over existing methods, particularly excelling on highly imbalanced datasets with up to 55\% F1-score improvements.
\end{abstract}
\begin{figure*}[t]
\centering
\includegraphics[width=\textwidth, trim={0.0cm 7.75cm 0.0cm 3cm}, clip]{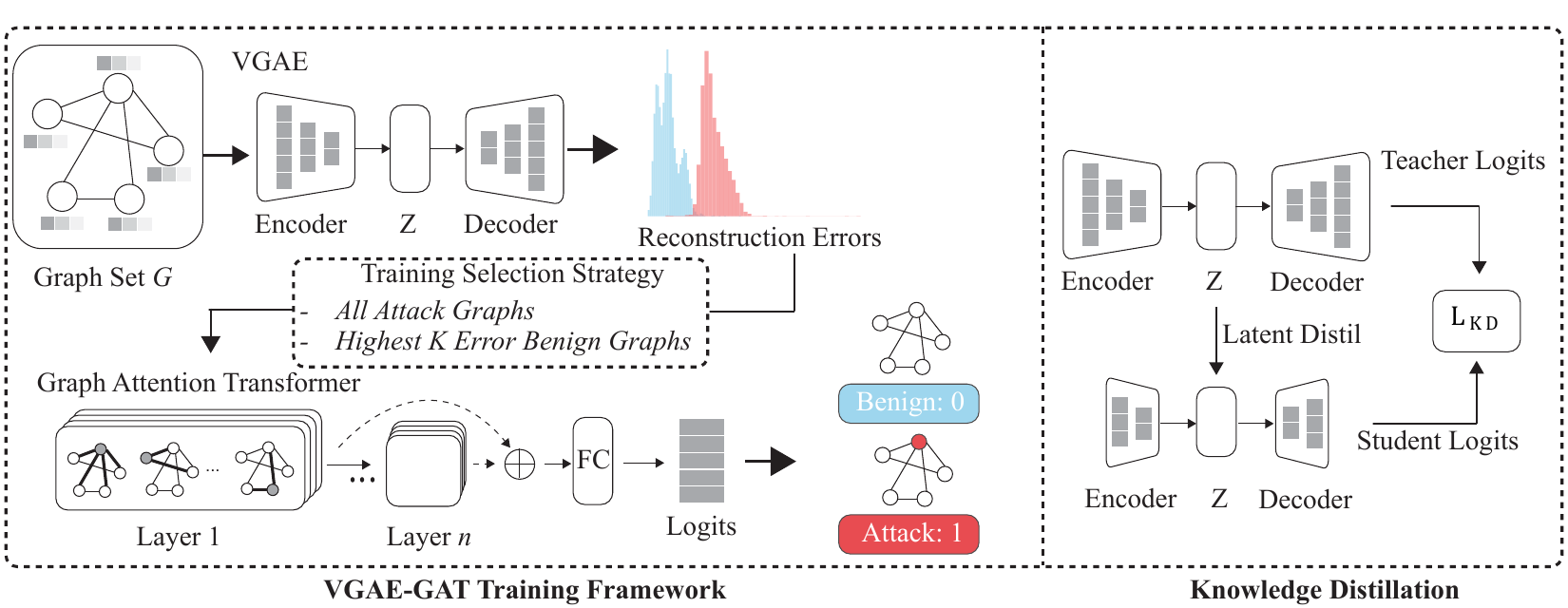}
\caption{Overview of the proposed multi-stage intrusion detection architecture. The left panel shows the two-stage framework: a Variational Graph Autoencoder (VGAE) performs anomaly scoring and selective undersampling, followed by a Graph Attention Network (GAT) for classification. $\bigoplus$ denotes concatenation of all GNN layer outputs before the final classifier. To mitigate class imbalance, the VGAE selects normal samples with highest reconstruction errors (hardest-to-classify examples) for GAT training at a 4:1 normal-to-attack ratio. The right panel illustrates knowledge distillation, where compact student models learn to replicate both predictions and latent representations of their corresponding teacher models. $n$ indicates the number of layers in each architecture.}
\label{framework}
\end{figure*}
\section{Introduction}

Modern vehicles rely on networks of electronic control units (ECUs) to manage everything from engine functions to advanced driver assistance systems (ADAS). Communication between ECUs is typically handled by the controller area network (CAN) protocol, valued for its reliability and cost-effectiveness in in-vehicle networks (IVNs)~\cite{Pazul}. However, CAN lacks built-in security mechanisms like encryption and authentication, as it was designed under the assumption of a closed, isolated network~\cite{ChoiK}.

With the introduction of on-board diagnostics (OBD) ports and wireless connectivity (e.g., Wi-Fi, cellular, V2X), access to the CAN bus has expanded significantly, opening new attack surfaces~\cite{Miller}. Attacks may now originate from both physical interfaces (OBD-II, USB) and remote channels (Bluetooth, mobile networks), allowing adversaries to inject malicious messages and potentially disrupt or take control of safety-critical vehicle systems~\cite{Woo, Wen}.

To counter these threats, intrusion detection systems (IDS) for CAN have become an area of active research. Traditional IDS approaches fall into two main categories: packet-based and window-based methods. Packet-based IDSs analyze individual CAN messages for quick detection, but cannot capture context or correlations across packets, limiting their effectiveness against complex attacks such as spoofing or replay~\cite{Cho}. Window-based IDSs consider sequences of packets, enabling better detection of such attack patterns, but often face challenges with detection delays and performance under low-volume or replay attacks~\cite{Muter, Han}.

Recent efforts address these limitations with statistical approaches using graph models~\cite{Islam}, advanced machine learning techniques such as deep convolutional neural networks (DCNN)~\cite{Kim}, and lightweight KNN classifiers~\cite{Derhab}. Other studies leverage temporal or dynamic graph features for high-accuracy detection of diverse attack types~\cite{Malik, Jiaru, AD}.

Despite strong results—for example, GNN- and VAE-based systems achieving over 97\% accuracy—adaptability and efficiency remain concerns for deployment in resource-constrained automotive environments~\cite{KD}. Knowledge distillation, explainable AI, and graph-based feature learning are promising strategies for developing robust, lightweight IDS that can operate effectively with limited resources and provide greater transparency.

\subsection{Motivation}
The CAN protocol is the backbone of in-vehicle communication but remains highly vulnerable to cyber-attacks. While a range of IDSs have been proposed, existing methodologies show critical limitations. Traditional packet-based IDSs, for example, largely ignore sequential context and are therefore ineffective in capturing correlations across consecutive packets thus restricting their ability to detect sophisticated attacks and to perform accurate attack classification. Window-based IDSs attempt to address temporal dependencies; however, this area remains nascent, with few frameworks achieving both high detection accuracy and lightweight deployment. In addition, the nature of this data introduces a pronounced class imbalance, with malicious activity being rare compared to benign traffic, often leading to biased detectors and poorly calibrated predictions.

To address these limitations, we propose a multistage graph neural network (GNN)-based framework that combines Variational Graph Autoencoders (VGAE) for unsupervised anomaly detection with Graph Attention Networks (GAT) for robust attack classification. Our approach captures structural dependencies in CAN traffic and uses KD to train a lightweight student model for efficient deployment on resource-constrained automotive platforms, while also employing targeted methods to mitigate class imbalance and improve rare attack detection.

\subsection{Contributions}
The main contributions of this research are as follows:
\begin{itemize}
    \item We propose a novel two-stage intrusion detection framework for CAN bus traffic that leverages both VGAE and GAT modules as shown in Figure \ref{framework}. In this architecture, VGAE serves as the initial stage for robust representation learning and anomaly score generation, while GAT performs refined classification by exploiting graph-structured dependencies.
    \item We introduce a KD strategy tailored for CAN bus applications on resource-limited edge devices, yielding a student model that is only 1.54\% the size of the teacher model but retains strong detection performance.
    \item We conduct comprehensive experiments on six publicly available CAN intrusion datasets, including full-sequence evaluation of the newly released can-train-and-test \cite{Lampe2024cantrainandtest} benchmark.
\end{itemize}

\section{Related Work}
Intrusion detection systems (IDS) for in-vehicle CAN networks are commonly classified by frame count, data type, and detection model~\cite{Dupont}. By frame count, IDSs can be divided into packet-based and window-based approaches. Packet-based IDSs analyze individual CAN frames for fast detection, but cannot capture dependencies between packets, which limits their accuracy. For example, Kang et al.~\cite{Kang} used deep neural networks on simulated data, and Groza et al.~\cite{Groza} applied Bloom filters to exploit traffic periodicity, though their method is ineffective for aperiodic frames~\cite{Cheng}.

Window-based IDSs instead analyze sequences of CAN frames, enabling temporal correlation analysis. Olufowobi et al.~\cite{Olufowobi} employed timing models for real-time detection without relying on specifications, but still struggled with aperiodic messages and repeated IDs. Frequency and Hamming distance methods~\cite{Taylor} are similarly less effective against aperiodic attacks~\cite{Bozdal,Choi}. Islam et al.~\cite{Islam} used graph features and statistical tests for anomaly and replay detection, but with increased detection latency due to the need for larger batches of messages.

Graph-based IDSs better capture ECU communication patterns but often target only simple attacks. G-IDCS~\cite{GIDCS} addresses this by combining an interpretable, threshold-based stage with a classifier leveraging message correlation, enabling detection of complex attacks beyond what packet-based IDSs can achieve.

Most CAN bus IDSs are anomaly-based, using rule-based detection for known signatures but struggling to generalize to novel attacks. To address this, recent work uses machine learning (ML) and deep learning (DL) to learn normal behavior and identify deviations. For instance, a CNN-LSTM-attention hybrid~\cite{CNNLSTM} achieved over 98

Recently, KD-GAT~\cite{frenken2025kdgat} combines Graph Attention Networks and knowledge distillation to obtain a lightweight yet effective IDS. The student model closely matches the teacher in accuracy but, like others, still struggles with severe class imbalance.

Despite progress, current IDSs face challenges including interpretability, scalability, and computation overhead, particularly for deployment in embedded automotive systems. Class imbalance and scarce labeled attack data further hinder real-world generalization. These challenges motivate hybrid and distillation approaches, such as combining GNNs and autoencoders with KD, as explored in this work.

\section{Background}
This section will cover fundamental concepts of the CAN protocol, GNNs, VGAE, and KD.
\subsection{CAN Protocol}
\begin{figure}[h] 
    \centering
\includegraphics[width=0.48\textwidth, trim={0.0cm 0cm 0.0cm 0cm}, clip]{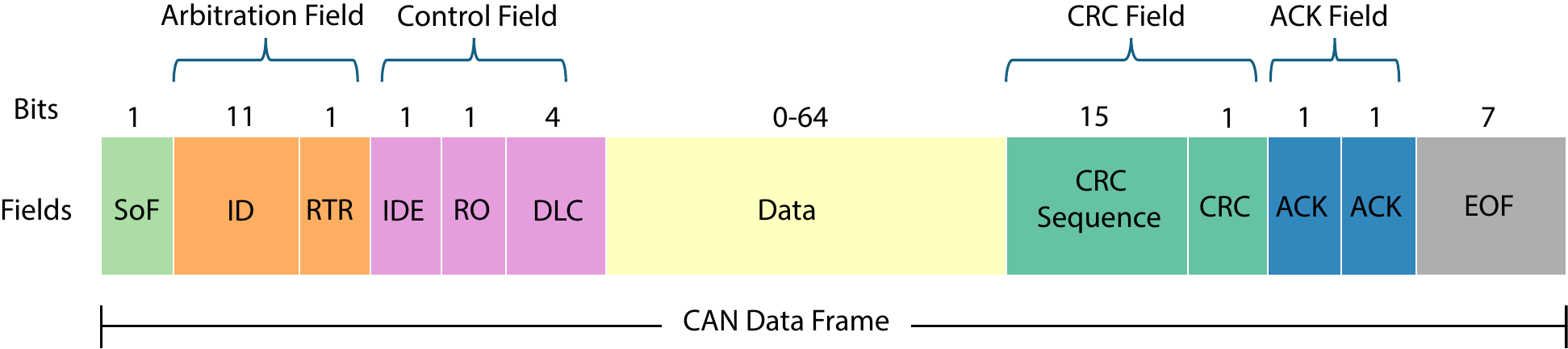}
   \caption{CAN data frame structure}
    \label{CAN_frame}
\end{figure}

The CAN is a robust serial protocol enabling real-time communication between electronic control units (ECUs) in vehicles. In a CAN bus, nodes broadcast messages, while receivers filter and process relevant ones. As shown in Figure \ref{CAN_frame}, each CAN data frame includes a Start-of-Frame, Arbitration, Control, Data, CRC, Acknowledgment, and End-of-Frame field. 

\subsection{Graph Neural Networks}
A graph is a data structure consisting of a set of nodes  \textit{V} and a set of edges \textit{E} that connect pairs of nodes. A graph can be defined as $G = (V,E)$, where $V = \{v_1, v_2, ..., v_n\}$ is a node set with $n$ nodes, and $E = \{e_1, e_2, ..., e_m\}$ is an edge set with $m$ edges. 
Given this graph structure, a GNN looks to find meaningful relationships and insights of the graph. The most common way to accomplish this is through the message passing framework \cite{gilmer2017mpnn} \cite{scarselli2009gnn}, where at each iteration, every node aggregates information from its local neighborhood. Across iterations, node embeddings contain information from further parts of the graph. This update rule can be explained through the following equation:
\begin{equation}
\mathbf{h}_v^{(k)} = \phi\big(\mathbf{h}_v^{(k-1)},\ \bigoplus_{u \in \mathcal{N}(v)} \psi(\mathbf{h}_v^{(k-1)}, \mathbf{h}_u^{(k-1)}, \mathbf{e}_{vu})\big)
\end{equation}

where $h$ is the feature embedding, $\phi$ is the node update function, $\psi$ the message function, $\mathbf{e}_{vu}$ the edge feature, $\bigoplus$ an aggregation (sum/mean), and $\mathcal{N}(v)$ the neighbors of $v$.

GAT \cite{velickovic2018gat} builds upon GNNs by introducing an attention mechanism. This allows each node in the message passing framework to dynamically assign weight contributions to their neighbors. For node $v$, the attention coefficient $\alpha_{vu}$ for neighbor $u$ is computed as:
\begin{equation}
\alpha_{vu} = \mathrm{softmax}\left(
    \mathrm{LeakyReLU}\left(
        \mathbf{a}^\top 
        \left[
            \mathbf{W}\mathbf{h}_v \mathbin{\|} \mathbf{W}\mathbf{h}_u
        \right]
    \right)
\right)
\end{equation}
where $\alpha$ is the learnable attention vector, $\mathbf{W}$ is a weight matrix, and $\mathbin{||}$ denotes concatenation between the weight matrices.

The attention function computes a scalar weight for each neighbor of node $v_i$, denoted by $\alpha_{ij}$, which reflects the importance or relevance of node $v_j$ for node $v_i$.
\begin{equation}
\mathbf{h}_v^{(k)} = \sigma\left(
    \sum_{u \in \mathcal{N}(v)} 
    \alpha_{vu} \mathbf{W} \mathbf{h}_u^{(k-1)}
\right)
\end{equation}
where $\sigma$ is the activation function, normally ELU or RELU.

The Jumping Knowledge (JK) module \cite{JumpingKnowledge} enhances GATs by aggregating intermediate layer representations. In this work, we adopt the concatenation strategy, where each node's final representation is formed by directly concatenating its embeddings from all GAT layers. Let $h_v^{(l)}$ denote the representation of node $v$ at layer $l \in {1, \dots, L}$. The final output is computed as:

\begin{equation}
\label{concat_jk}
h_v^{\text{final}} = \left[ h_v^{(1)} , ; , h_v^{(2)} , ; , \dots , ; , h_v^{(L)} \right]
\end{equation}
This approach preserves multi-level features without introducing additional sequential modeling overhead.

\subsection{Variational Graph Autoencoder}
The Variational Graph Autoencoder (VGAE)~\cite{kipf2016variational} is a probabilistic model designed for unsupervised learning on graphs. Given a graph $G=(V, E)$ with adjacency matrix $A$ and node features $X$, VGAE approximates the posterior distribution of latent variables $Z$ using a two-layer GCN encoder.

The encoder approximates the posterior distribution over the latent variables \( Z = \{z_1, ..., z_N\} \) by assuming a Gaussian distribution for each node:
\begin{equation}
    q(Z|X, A) = \prod_{i=1}^{N} \mathcal{N}(z_i|\mu_i, \mathrm{diag}(\sigma_i^2)),
\end{equation}
where \( \mu_i \in \mathbb{R}^d \) and \( \sigma_i \in \mathbb{R}^d \) are the mean and standard deviation vectors for node \( i \). These are parameterized by two separate GCN layers:
\[
\mu = \mathrm{GCN}_\mu(X, A), \quad \log \sigma = \mathrm{GCN}_\sigma(X, A),
\]
which capture both local topology and node features. The outputs of these GCNs define the variational posterior \( q(Z|X, A) \).

The decoder attempts to reconstruct the graph structure by computing the probability of edge existence between any two nodes \( i \) and \( j \) as:
\begin{equation}
    p(A|Z) = \prod_{i=1}^{N} \prod_{j=1}^{N} \sigma(z_i^\top z_j),
\end{equation}
where \( \sigma(\cdot) \) is the sigmoid function and \( z_i^\top z_j \) measures similarity in latent space. This inner product decoder encourages connected nodes to have similar embeddings.

The training objective is to maximize the variational evidence lower bound (ELBO), which consists of a reconstruction term and a regularization term:
\begin{equation}
    \mathcal{L} = \mathbb{E}_{q(Z|X, A)}[\log p(A|Z)] - \mathrm{KL}[q(Z|X, A) || p(Z)],
\end{equation}
where the first term encourages accurate reconstruction of the observed adjacency matrix, and the second term is the Kullback-Leibler divergence between the approximate posterior and the prior \( p(Z) = \prod_{i=1}^N \mathcal{N}(z_i|0, I) \), promoting regularization and disentangled latent representations.

While VGAE effectively captures global graph structure, its full-graph decoding may be suboptimal for detecting localized anomalies, especially in sparse or noisy graphs. To address this, Zhou et al.~\cite{zhou2023gadnr} introduced GAD-NR, which replaces full adjacency reconstruction with localized neighborhood prediction. This modification enhances sensitivity to topological deviations at the node-level, making it suitable for intrusion detection in systems like CAN networks. Inspired by this, our architecture adopts neighborhood-level reconstruction via masked decoding over the graph of each CAN window.
\subsection{Knowledge Distillation}
KD, popularized by Hinton et al.\cite{hinton2015distilling}, is a widely adopted model compression technique where a small, efficient student model is trained to reproduce the behavior of a large, accurate teacher model. 
The soft target probabilities output by a teacher model encode rich relational information between classes that's often not captured by hard labels alone. Training a student model to match these softened outputs enables it to learn a more informative function approximation than training with one-hot labels alone.

Concretely, given an input $x$, the teacher produces a vector of logits $s^t(x)$, which are converted into a softened distribution via temperature scaling $\tau$:
\begin{equation}
\tilde{p}^t_k(x) = \frac{\exp(s^t_k(x)/\tau)}{\sum_j \exp(s^t_j(x)/\tau)}
\end{equation}

The student is trained to match these probabilities by minimizing the Kullback-Leibler divergence between teacher and student distributions (distillation loss), alongside the standard supervised classification loss:
\begin{equation}
\mathcal{L}_{\text{total}} = \alpha \cdot \mathcal{L}_{\text{hard}} + (1 - \alpha) \cdot \mathcal{L}_{\text{KD}}
\end{equation}

where $\alpha$ balances the contribution of ground truth and teacher supervision.
\section{Methodology} \label{methodology}
The two-stage framework depicted in Figure \ref{framework} is explained in detail in this section.
\subsection{Graph Construction}

CAN data is typically represented as a tabular time series dataset, where each message \( m_i \) is characterized by attributes such as CAN ID, payload data, and timestamp. To convert this sequential data into graph inputs suitable for graph-based intrusion detection, we define the following procedure:

\paragraph{Sliding Window and Graph Formation}  
Given a sliding window of size \( W \) (set to 100 messages in this work), we extract a subsequence of CAN messages:
\[
\mathcal{W}_t = \{ m_{t}, m_{t+1}, \ldots, m_{t+W-1} \}
\]
where each message \( m_i = ( \text{ID}_i, \text{payload}_i ) \).

\paragraph{Node Definition}  
Each unique CAN ID within the window \(\mathcal{W}_t\) corresponds to a node \( v_j \) in the graph \( G_t = (V_t, E_t) \). The node attributes are defined as:
\begin{equation}
\mathbf{x}_j = \left[
\text{ID}_j,
f_j = \frac{\text{count}(\text{ID}_j)}{W},
\bar{p}_j = \frac{1}{n_j} \sum_{k=1}^{n_j} \text{payload}_{j,k}
\right]
\end{equation}
where \( n_j \) is the number of occurrences of CAN ID \( j \) in the window, and \(\bar{p}_j\) is the average payload value for that ID.

\paragraph{Edge Construction}  
Edges \( e_{jk} \in E_t \) are created between nodes \( v_j \) and \( v_k \) if their corresponding messages appear sequentially in the window. Formally, for message pairs \((m_i, m_{i+1})\), if \( m_i \) corresponds to node \( v_j \) and \( m_{i+1} \) to node \( v_k \), then:
\[
e_{jk} = \text{number of occurrences of } (v_j, v_k) \text{ in } \mathcal{W}_t
\]
This captures the temporal relational structure of CAN messages within the window.

\begin{figure}[h]
        \centering
        \includegraphics[width=\linewidth, trim={0.5cm 0.20cm 0.5cm 0.60cm}, clip]{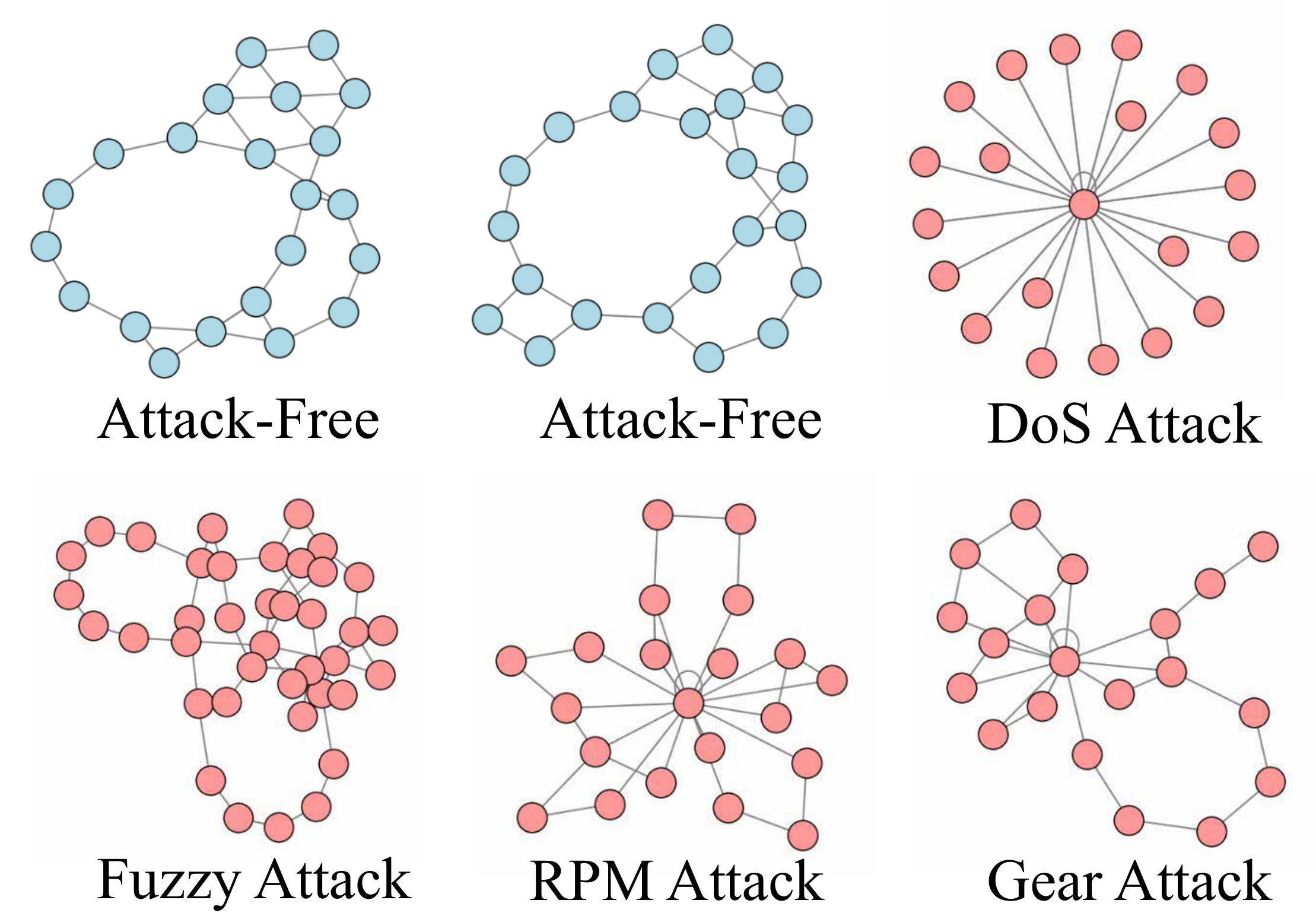}
        \caption{Graphs created from the Car Hacking Dataset. Each node represents a unique CAN ID found in the window, and each edge is constructed between sequential IDs, including self edges. The feature vector associated with each node is not shown. Blue denotes that the graph has no attack IDs, while red indicates that at least one of the nodes is an intrusion.}
        \label{fig:graph-examples}
    \end{figure}

\paragraph{Graph Input Summary}  
Thus, each sliding window \(\mathcal{W}_t\) is represented as a graph \( G_t = (V_t, E_t, X_t) \) where \(V_t\) are nodes (unique CAN IDs), \(E_t\) are edges (sequential message relations), and \(X_t\) are node attributes as defined above. Labels are created for a binary classification task, where $0$ if $W_t$ contains only benign messages, and $1$ if $W_t$ contains any attack messages. Figure \ref{fig:graph-examples} shows some examples of attack-free and attack graphs.

\subsection{Training Paradigm}
Our approach employs a novel two-stage training framework that leverages the complementary strengths of VGAE and GAT for enhanced graph-based classification while addressing class imbalance through selective undersampling and knowledge distillation.

\textbf{Stage 1: VGAE Training and Selective Undersampling.} The first stage trains a VGAE model exclusively on normal graph samples to learn baseline structural patterns. The VGAE encoder generates latent representations $Z \sim \mathcal{N}(\mu, \sigma^2)$ from input graphs, while the decoder reconstructs adjacency matrices. Following training, we implement selective undersampling based solely on reconstruction errors: $R_{error}(i) = ||A_i - \hat{A}_i||_2$ for each normal graph sample. This identifies normal samples with highest reconstruction errors—those most difficult to reconstruct and likely on decision boundaries—maintaining a 4:1 normal-to-attack ratio for Stage 2.

\textbf{Stage 2: GAT Training with Enhanced Representations.} The second stage trains a GAT on the filtered dataset using multi-head attention mechanisms: $\alpha_{ij}^{(l,h)} = \text{softmax}_j(e_{ij}^{(l,h)})$ to learn adaptive neighborhood weights for flexible neighborhood aggregation.

\textbf{KD Training:} Both stages can be enhanced through knowledge distillation, where the entire two-stage framework is re-executed with student models learning from pre-trained teacher models. In this paradigm, student models receive guidance through soft labels: $\mathcal{L}_{KD} = \text{KL}(P_{student}^{soft} || P_{teacher})$, where teacher predictions are softened using temperature scaling. For VGAE, knowledge transfer occurs in the latent space, while GAT distillation operates on final node representations. This approach improves generalization and convergence by leveraging the learned expertise of teacher models.

\section{Experiments}
This section presents the experimental setup, evaluation metrics, training details, and insights into the datasets used in this study.

\subsection{Experimental Setup}
Table \ref{tab:teacher_student_multi} highlights the differences between the teacher and student model for both the VGAE and GAT models.
\begin{table}[h]
    \centering
    \caption{Teacher (Tch.) and Student (Std.) Model Parameters for Autoencoder and Classifier}
    \begin{tabular}{lcccc}
        \toprule
        & \multicolumn{2}{c}{\textbf{Autoencoder}} & \multicolumn{2}{c}{\textbf{Classifier}} \\
        \cmidrule(lr){2-3} \cmidrule(lr){4-5}
        \textbf{Param.} & \textbf{Tch.} & \textbf{Std.} & \textbf{Tch.} & \textbf{Std.} \\
        \midrule
        Layers         & 3   & 2   & 5     & 2     \\
        Attn. heads        & 4   & 2   & 8     & 4     \\
        Hidden ch.         & 32   & 16   & 32    & 16    \\
        Loss fn.           & BCE   & KL Div. & BCE   & KL Div. \\
        Tot. param.        & 184K  & 87K  & 3.56M  & 55K  \\
        \bottomrule
    \end{tabular}
    \label{tab:teacher_student_multi}
\end{table}

\begin{table*}[h]
\centering
\caption{Test Set Performance across CAN Intrusion Datasets}
\setlength{\tabcolsep}{7pt}
\begin{tabular}{l*{2}{r}|*{2}{r}|*{2}{r}|*{2}{r}|*{2}{r}|*{2}{r}}
\toprule
 & \multicolumn{2}{c|}{CarH} & \multicolumn{2}{c|}{CarS} & \multicolumn{2}{c|}{S01} & \multicolumn{2}{c|}{S02} & \multicolumn{2}{c|}{S03} & \multicolumn{2}{c}{S04} \\
\cmidrule(lr){2-3} \cmidrule(lr){4-5} \cmidrule(lr){6-7} \cmidrule(lr){8-9} \cmidrule(lr){10-11} \cmidrule(l){12-13}
Method & Acc. & F1 & Acc. & F1 & Acc. & F1 & Acc. & F1 & Acc. & F1 & Acc. & F1 \\
\midrule    
A\&D       & 99.95 & 99.94 & -- & -- & -- & -- & -- & -- & -- & -- & -- & -- \\
G-IDCS     & 97.25 & 93.36 & -- & -- & -- & -- & -- & -- & -- & -- & -- & -- \\
GUARD-CAN      & 97.02 & 97.27 & -- & -- & -- & -- & -- & -- & -- & -- & -- & -- \\
KD-GAT & 99.97 & 99.97 & 99.31 & 99.29 & 99.29 & 88.08 & 98.18 & 24.42 & 98.24 & 86.06 & 87.07 & 61.35 \\
\textbf{Ours} & \textbf{99.89} & \textbf {99.89} & \textbf{99.96} & \textbf {99.96} & \textbf{99.38} & \textbf{89.86} & \textbf{ 99.61} & \textbf{79.67} & \textbf{99.29} & \textbf{95.10} &\textbf{ 96.47} & \textbf{91.99} \\
\bottomrule
\end{tabular}
\label{tab:results_metrics}
\end{table*}
\textbf{Evaluation Metrics}
The performance of the model is evaluated using accuracy and F1-Score.

\textbf{Implementation Details}
80\% of the dataset was utilized for training, 20\% for validation, and a distinct test set was used compiled by the dataset providers. Include batch size, learning rate, and epochs. All experiments were conducted using PyTorch and PyTorch Geometric. Model training and evaluation were performed on GPU clusters provided by the Ohio Supercomputer Center (OSC)\cite{OhioSupercomputerCenter1987}.

\subsection{Datasets}
Our proposed method has been evaluated on three publicly available automotive CAN intrusion detection datasets, each offering distinct characteristics and challenges for comprehensive IDS evaluation. Below provides a detailed comparison of dataset specifications and characteristics.

\textbf{1. HCRL Car-Hacking:} This dataset contains CAN traffic from a Hyundai YF Sonata with four attack types: DoS, fuzzing, RPM spoofing, and gear spoofing. All attacks were conducted on a real vehicle, with data logged via the OBD-II port. The dataset includes 988,872 attack-free samples and approximately 16.6 million total samples across all attack types\cite{Song2020carhacking}.

\textbf{2. HCRL Survival Analysis:} Collected from three vehicles (Chevrolet Spark, Hyundai YF Sonata, Kia Soul), this dataset enables scenario-based evaluation with three attack types: flooding (DoS), fuzzing, and malfunction (spoofing). The dataset is structured with 627,264 training samples and four testing subsets designed to evaluate IDS performance across known/unknown vehicles and known/unknown attacks \cite{Han2018survival}.

\textbf{3. can-train-and-test:}\footnote{https://bitbucket.org/brooke-lampe/can-train-and-test-v1.5/src/master/} The largest dataset, containing CAN traffic from four vehicles across two manufacturers (GM and Subaru). It provides nine distinct attack scenarios including DoS, fuzzing, systematic, various spoofing attacks, standstill, and interval attacks. The dataset is organized into four vehicle sets (set\_01 to set\_04) with over 192 million total samples. This dataset exhibits extreme class imbalance with attack-free to attack sample ratios ranging from 36:1 to 927:1 across different subsets. Each set contains one training subset and four testing subsets following the known/unknown vehicle and attack paradigm\cite{Lampe2024cantrainandtest}. This work will limit evaluation to the known vehicle and attack testing set.

\section{Results and Discussion}
\subsection{Experimental Results}

Table~\ref{tab:results_metrics} summarizes the test set performance across six datasets. We compare against four GNN-based baselines: KD-GAT~\cite{frenken2025kdgat}, A\&D GAT~\cite{AD}, G-IDCS~\cite{Park}, and GUARD-CAN~\cite{guardcan2025}. KD-GAT serves as the primary baseline since it is the only method evaluated on the comprehensive can-train-and-test dataset~\cite{Lampe2024cantrainandtest}.

Our approach demonstrates consistent improvements across all datasets, with particularly significant gains on highly imbalanced datasets. Compared to KD-GAT, we achieve an average improvement of 2.09\% in accuracy and 16.22\% in F1-score. The most substantial improvements occur on challenging datasets S02 and S04, where F1-scores improve by 55.25\% and 30.64\% respectively, indicating superior handling of severe class imbalance.

\subsection{Discussion}
\textbf{Class Imbalance Handling:} Our multi-stage approach demonstrates superior performance on imbalanced datasets compared to single-stage methods. The VGAE component effectively captures structural anomalies even with limited attack samples, while the GAT classifier benefits from the refined feature representations. This combination proves particularly effective on datasets S02 and S04, where traditional methods struggle with extreme class ratios.

\textbf{Generalization Capability:} The consistent performance across diverse datasets (CarH, CarS, and can-train-and-test subsets) demonstrates strong generalization. Unlike previous methods that show significant performance degradation on unseen test data, our approach maintains robust detection capabilities across different attack types and network conditions.


\textbf{Limitations:} While our method shows substantial improvements, performance on extremely imbalanced datasets (e.g., S02 with 1.14\% attack samples) remains challenging for the entire field. Future work should explore advanced sampling strategies and loss functions specifically designed for such scenarios.

\section{Ablation Study}
To assess the contribution of each model component, we perform ablation experiments comparing standalone and fused architectures. Table~\ref{tab:ablation_compact} summarizes F1-scores for the GAT-only and fusion setups across all datasets.

\textbf{Score Fusion:} During inference, predictions are fused using fixed performance-based weights as $P_{\text{fused}} = \omega_{\text{anomaly}} \cdot P_{\text{VGAE}} + \omega_{\text{GAT}} \cdot P_{\text{GAT}}$, where $\omega_{\text{anomaly}} = 0.15$ and $\omega_{\text{GAT}} = 0.85$. These weights (0.85, 0.15) were determined empirically based on validation performance.

The ablation results show that GAT-only performs best or on par with the weighted fusion approach across all datasets, suggesting that simple linear fusion offers limited benefits over the standalone GAT classifier. Future work should explore more sophisticated fusion strategies, such as mixture-of-experts (MoE) architectures or attention-based fusion mechanisms, to better leverage the complementary strengths of both components.
\begin{figure*}[h]
    \centering
    \begin{subfigure}{0.44\textwidth}
        \includegraphics[width=\linewidth]{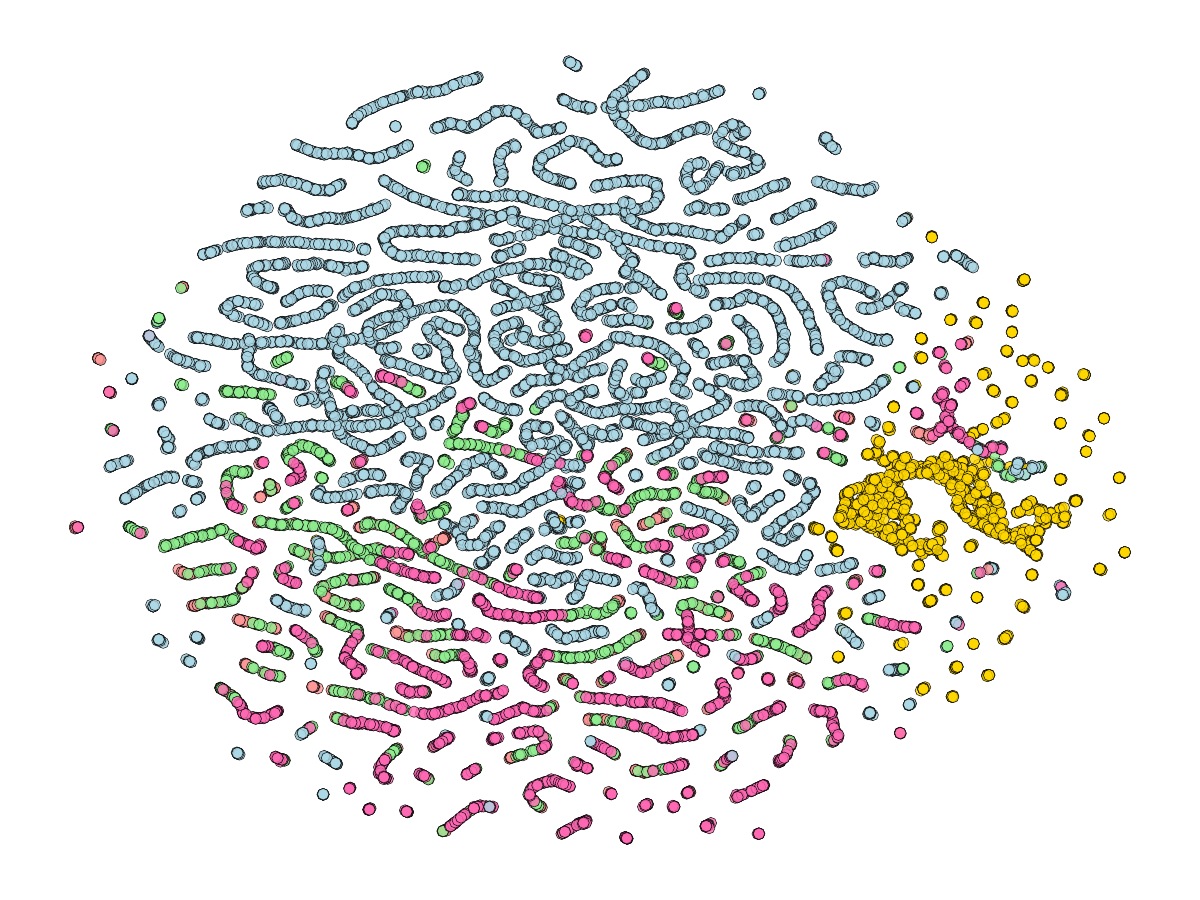}
        \caption{UMAP of raw CAN messages}
        \label{fig:umap-raw}
    \end{subfigure}
    \hfill
    \begin{subfigure}{0.44\textwidth}
        \includegraphics[width=\linewidth]{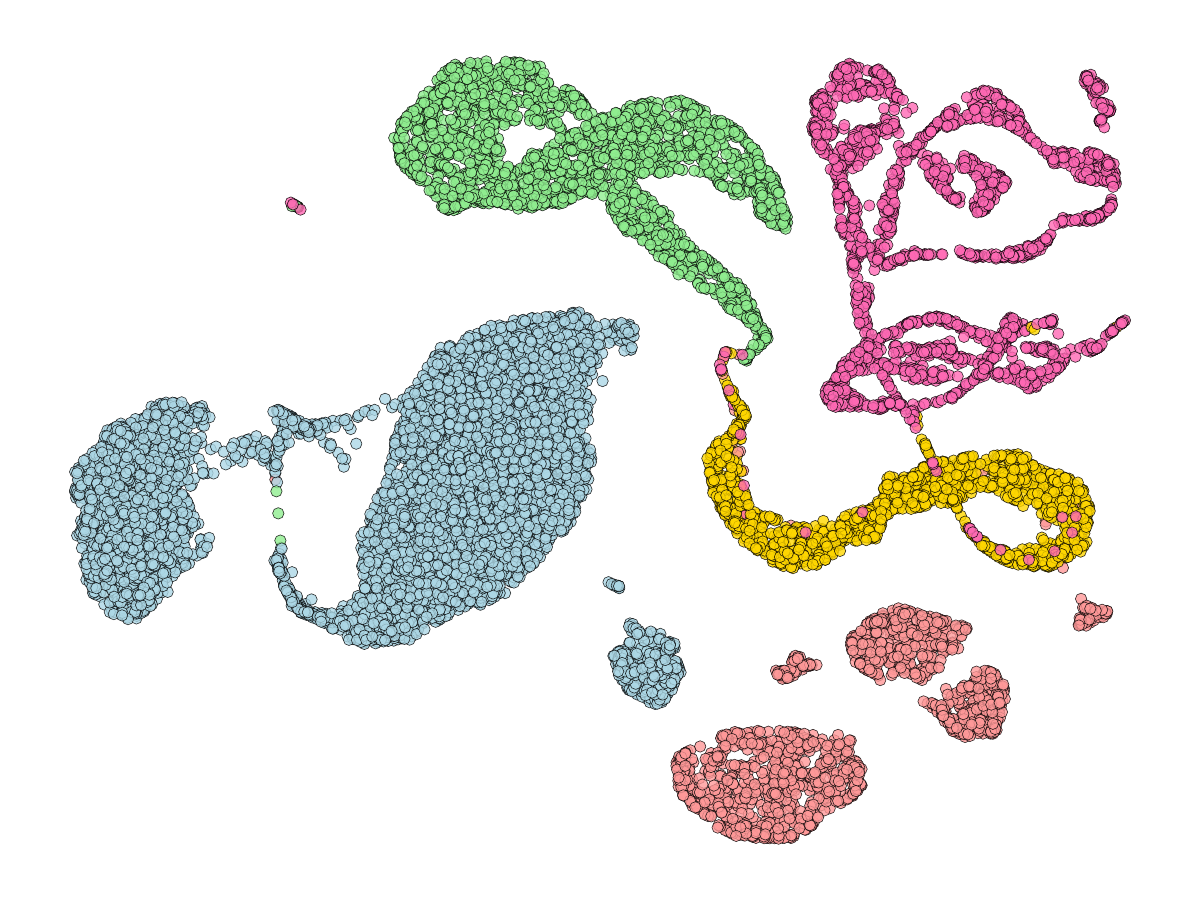}
        \caption{UMAP of GAT graph-level embeddings}
        \label{fig:umap-gat}
    \end{subfigure}
    \includegraphics[width=0.6\linewidth, trim={0.0cm 0.5cm 0.0cm 0.5cm}, clip]{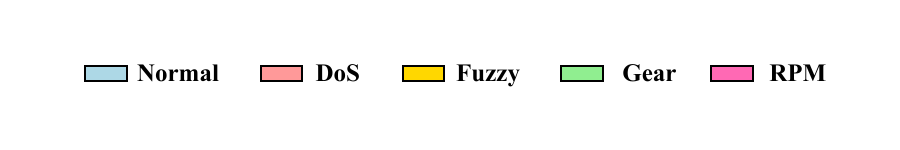}
    \caption{
        UMAP visualizations of sampled CAN messages (left) and learned GAT embeddings (right). Even though the model was trained using binary labels, it naturally separates different attack types in the learned feature space.
    }
    \label{fig:umap-combined}
\end{figure*}

\begin{figure}[h]
        \centering
        \includegraphics[width=\linewidth, trim={0.0cm 0.0cm 0.0cm 0.0cm}, clip]{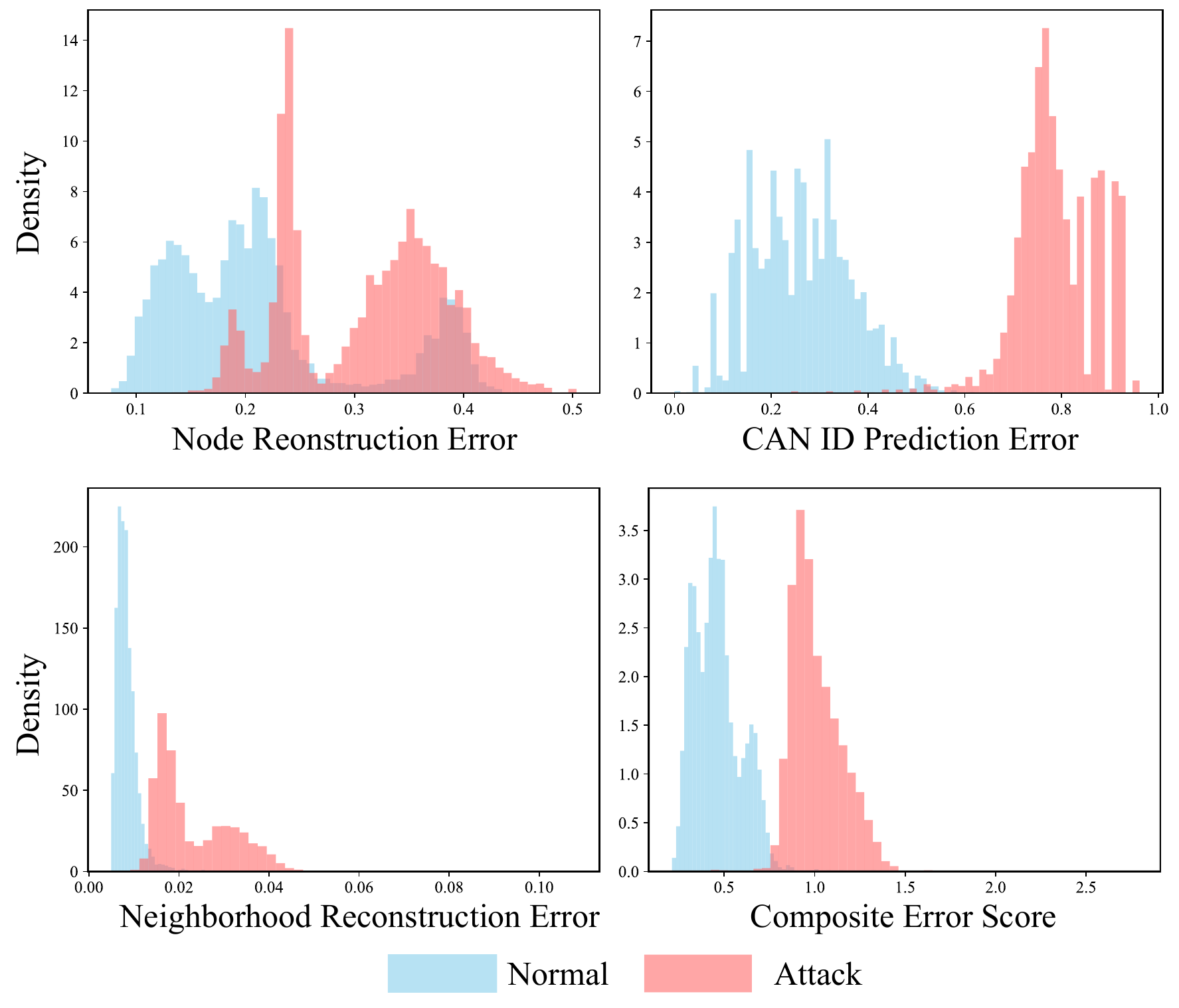}
        \caption{Reconstruction Error of VGAE.}
        \label{fig:VGAE}
    \end{figure}

\begin{table}[h]
\centering
\caption{Ablation Study Results (F1-Scores)}
\label{tab:ablation_compact}
\begin{tabular}{lccc}
\toprule
\textbf{Dataset} & \textbf{GAT-Only} & \textbf{Fusion} & \textbf{Best} \\
\midrule
S01 & \textbf{0.899} & 0.895 & GAT \\
S02 & \textbf{0.797} & 0.792 & GAT \\
S03 & 0.951 & 0.951 & Tie \\
S04 & \textbf{0.920} & 0.918 & GAT \\
CarH & 0.999 & 0.999 & Tie \\
CarS & 1.000 & 1.000 & Tie \\
\midrule
\textbf{Mean} & \textbf{0.927} & \textbf{0.926} & -- \\
\bottomrule
\end{tabular}
\end{table}
\section{Explainability}
\subsection{UMAP Analysis}
To understand the representations learned by our model, we perform UMAP-based feature analysis using both raw input statistics and learned graph embeddings. We sample 10\% of graphs from the HCRL Car-Hacking dataset.

Figure~\ref{fig:umap-combined}a visualizes processed CAN-graph data projected via UMAP. The loose clustering indicates limited separability between normal and attack types. In contrast, Figure~\ref{fig:umap-combined}b shows UMAP projections of graph-level embeddings from the trained GAT classifier's penultimate layer.

Despite binary supervision (attack vs. normal), the learned embedding space forms well-separated clusters aligned with specific attack types (DoS, Fuzzy, Gear, RPM). This emergent multi-class structure demonstrates that our model captures high-level semantic patterns in CAN traffic and generalizes across attack categories without explicit multi-class labels. The clear cluster separation in embedding space, absent in raw features, validates the GAT's ability to learn discriminative representations from graph-structured temporal data.

\subsection{Composite VGAE Reconstruction Error}
To assess the overall reconstruction quality of the VGAE, we combine three types of reconstruction errors: node feature reconstruction error (\( E_{\text{node}} \)), CAN ID prediction error (\( E_{\text{CAN\,ID}} \)), and neighborhood reconstruction error (\( E_{\text{neighbor}} \)). Each error captures a different aspect of the graph structure and message semantics. We compute a single composite score as a weighted sum:
\begin{equation}
\mathrm{Composite\_Error} = \alpha\, E_{\text{node}} + \beta\, E_{\text{neighbor}} + \gamma\, E_{\text{CAN\,ID}}
\end{equation}
where $ \alpha $, $ \beta $, and $ \gamma $ are empirically chosen weights that regulate each term's influence. In our experiments, we use $ \alpha = 1.0 $, $ \beta = 20.0 $, and $ \gamma = 0.3 $.
Figure \ref{fig:VGAE} shows the distribution of the individual error components and the resulting composite anomaly score. This approach enables the detection of subtle anomalies by jointly evaluating node content, CAN identifier semantics, and local neighborhood structure.

\section{Conclusion}
We introduced a novel multi-stage CAN intrusion detection framework combining Variational Graph Autoencoder and Graph Attention Network modules for robust anomaly detection and classification. Knowledge distillation enables a compact student model achieving 96\% parameter reduction while maintaining strong performance. Extensive experiments across six benchmark datasets demonstrate significant improvements over existing methods, with average F1-score gains of 16.2\% and exceptional performance on class-imbalanced scenarios. Our ablation study reveals that the standalone GAT classifier achieves comparable performance to fusion approaches with greater computational efficiency, making it ideal for resource-constrained automotive environments. These results highlight the promise of graph-based, multi-stage deep learning combined with knowledge distillation for practical automotive cybersecurity deployment.

\bibliography{aaai2026}

\end{document}